\documentclass[letterpaper]{article}
\usepackage{aaai}
\usepackage{times}
\usepackage{helvet}
\usepackage{courier}
\usepackage{graphicx}
\usepackage{hyperref}
\usepackage{booktabs}
\usepackage{amsmath}
\usepackage{tcolorbox} 
\usepackage{xcolor} 
\begin{document}
%
\title{Eyes on the Streets: Leveraging Street-Level Imaging to Model Urban Crime Dynamics}
\author{Zhixuan Qi\thanks{These authors contributed equally to this work.} \and Huaiying Luo\footnotemark[1] \and Chen Chi\\
Cornell Tech, New York, United States\\
\{zq83, hl2446, cc2787\}@cornell.edu
}

\definecolor{mybackground}{RGB}{255,253,208} 
\definecolor{myfontcolor}{RGB}{153,0,0} 

\tcbset{
    colback=mybackground,    
    colframe=mybackground,   
    coltext=myfontcolor,     
    boxrule=0pt,             
    arc=0pt,                 
    left=10pt,
    right=10pt,
    boxsep=5pt,
}

\maketitle
\begin{tcolorbox}
\textbf{Disclaimer:} This project employed crime rate data as a proxy metric, which we later recognized as problematic due to inherent biases and its inadequacy in holistically assessing community safety and well-being. Reported crime statistics often disproportionately stigmatize minority groups and fail to capture the factors influencing safety. Evaluating urban environments requires a more comprehensive, equitable, and community-driven approach that moves beyond an over-reliance on reported crime rates alone.
\end{tcolorbox}
\begin{abstract}
\begin{quote}
This study addresses the challenge of urban safety in New York City by examining the relationship between the built environment and crime rates using machine learning and a comprehensive dataset of street view images. We aim to identify how urban landscapes correlate with crime statistics, focusing on the characteristics of street views and their association with crime rates. The findings offer insights for urban planning and crime prevention, highlighting the potential of environmental design in enhancing public safety.
\end{quote}
\end{abstract}

\section{Introduction}
Safety in urban areas is a critical concern, particularly in densely populated metropolises like New York City, where the well-being of a vast community is at stake. The high crime rate has remained an unsettling challenge for residents across America's large cities. In response, various map applications have emerged, aimed at guiding citizens away from potentially perilous neighborhoods. Backed by environmental psychology, a wealth of research underscores the profound influence of the built environment features on human behavior. As a team of urban enthusiasts, we are motivated to explore how the built environment interplays with crime rates.

In this project, we aim to apply computer vision models to a comprehensive street view data of New York City, to understand the connection between the varying visages of city streets and the crime statistics tied to them. The specific goals of this study are to: 
\begin{itemize}
    \item Examine how street view characteristics are associated with crime rates, types and severity;
    \item Explore the implications of the findings for future urban environment interventions and crime prevention efforts.
\end{itemize}
The implications of our findings could be far-reaching, offering valuable insights for city officials, urban planners, NYPD, and residents in their quest to foster a safer living environment in New York City. This, eventually, could significantly bolster our ability to craft and implement effective urban design and management strategies, making cities not just infrastructural entities but safe havens for their inhabitants.

\section{Datasets}
\subsection{Street View}
Our baseline utilizes the Google Street View dataset provided by the University of Central Florida (UCF)\cite{zamir_image_2014}. This dataset contains a collection of 62,058 high-resolution images from Google Street View, showcasing urban scenes from Pittsburgh, PA; Orlando, FL; and parts of Manhattan, NY. Each image is paired with precise GPS data and compass orientation for accuracy. The data gathering spanned from February 13, 2010, to October 20, 2010.

\begin{figure}[ht]
    \centering
    \includegraphics[width=0.8\linewidth]{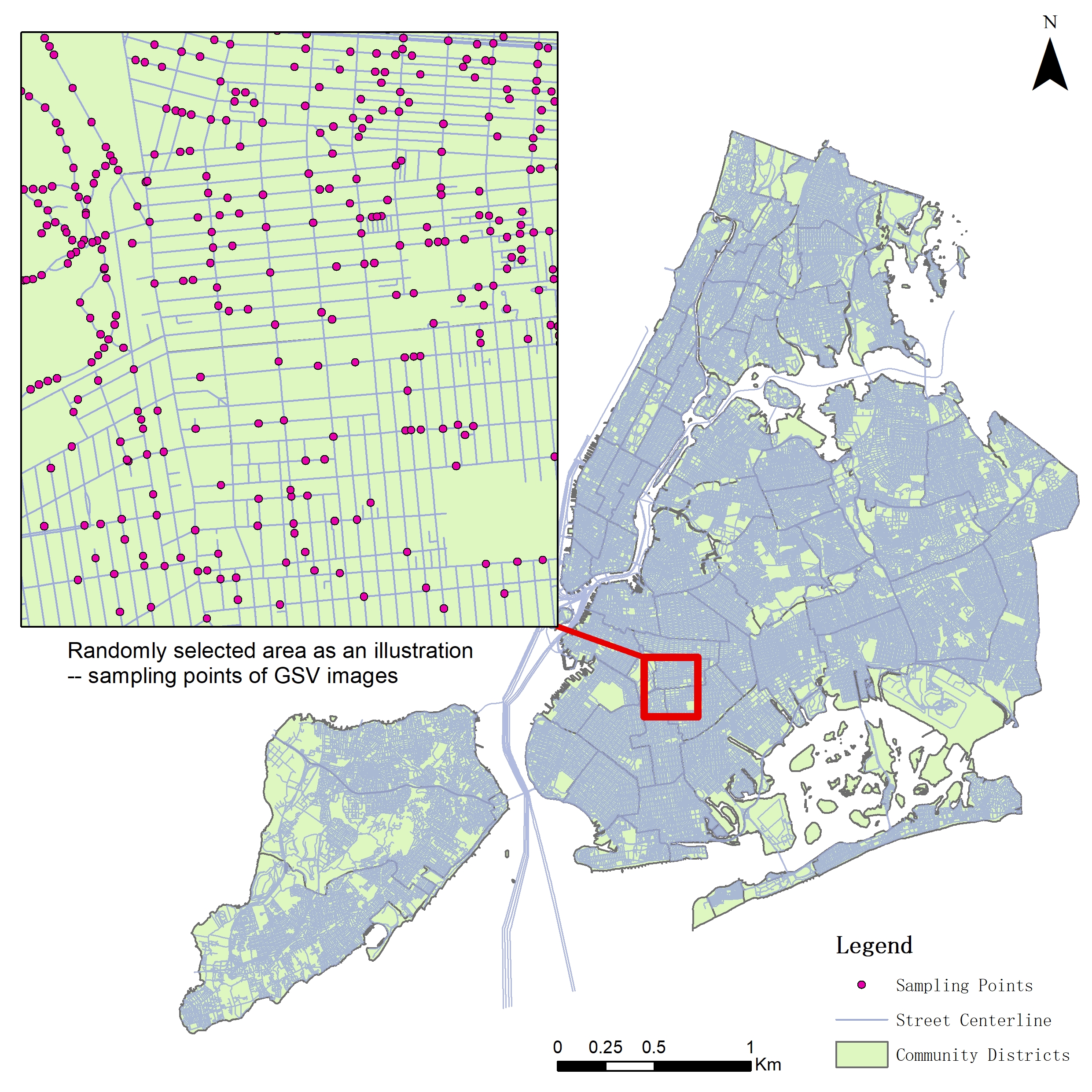}
    \caption{Sample points from the network.}
    \label{fig:sample_pts}
\end{figure}

For the primary component of our study, we independently constructed a dataset of street-view images. To ensure a diverse and representative portrayal of the street landscapes in New York City, we conducted a systematic analysis, selecting a point every 50 meters along the road centerlines throughout the city. Due to computational limitations and budgetary considerations, we randomly selected 200 points within each community to serve as our sampling points. For each of these points, we utilized the Google Static View API to request a street view image of dimensions 600x300 pixels. This API provided us with the most recent image available, ideally captured around the year 2022. Through this process, we amassed a total of 14,200 sampling points and successfully retrieved 13,636 images for 71 communities. Figure \ref{fig:sample_pts} shows the sampling points across the New York City street network.

\subsection{Crime Rate}
We obtain crime data for various locations within New York City from the dataset NYPD
Arrests Data (Historic) provided by the NYC Open Data Platform. This dataset catalogs all arrests made by the New York City Police Department (NYPD) from 2006 up to April 27, 2023. This extensive dataset is compiled through a meticulous process where data is extracted quarterly and undergoes a thorough review by the NYPD's Office of Management Analysis and Planning. It is then published on the NYPD's official website. Each entry in the dataset details an individual arrest,including information about the type of crime, the location and time of enforcement. It also includes demographic information about the suspects.

In this study, we focused on analyzing arrest data from the entire year of 2022, as this represents the most recent complete year of statistical data available to us. To assess the crime rate within individual communities, we merged the points of interest (POIs) related to crimes with community boundary data. This process enabled us to calculate and compare the crime rates for each community for the year 2022.

\subsubsection{Geographical Units}
Our analysis is structured around the community as the geographical unit. We source the demarcation of community boundaries from the NYC Open Data Platform, accessible at: \href{https://data.cityofnewyork.us/City-Government/Community-Districts/yfnk-k7r4}{Community Districts Dataset}.

Given that each entry in both the crime records and the street view image dataset includes coordinate information, we compile the crime rates and visual features of the street images according to each community. Through this aggregation, we've consolidated the information into 71 distinct data points. These compiled data points are then utilized to train and evaluate our predictive model, with the crime rates as the target variable. Table \ref{tab:community_stats} shows the descriptive statistics of crime variables, street tree measures, and socioeconomic variables by community.

\begin{table*}[ht]
\centering
\caption{Descriptive Statistics of Community Areas}
\label{tab:community_stats}
\begin{tabular}{@{}l l r r r r@{}}
\toprule
Variable & Description & \multicolumn{1}{c}{Mean} & \multicolumn{1}{c}{Stdv} & \multicolumn{1}{c}{Min} & \multicolumn{1}{c}{Max} \\
\midrule
Community Area (km\textsuperscript{2}) & Total area & 11.01 & 10.50 & 2.25 & 55.73 \\
Road Length (km) & Total road length & 177.60 & 146.43 & 19.08 & 696.57 \\
Crime Count & Total reported crimes & 2673.82 & 1798.62 & 2.00 & 7778.00 \\
Crime Rate & Rate of crimes & 0.01409 & 0.00947 & 0.00001 & 0.04097 \\
\bottomrule
\end{tabular}
\end{table*}

\section{Model}
\subsection{Semantic Segmentation}
Semantic Segmentation in urban analysis has evolved significantly with the advent of deep learning technologies. Its adaptability is highlighted in various studies, from enhanced street view analysis to applications in medical imaging. These models are also tested in urban settings and are considered to achieve good performance. For this study, we implemented and fine-tuned a DeepLabv3 for the segmentation task \cite{chen_rethinking_2017}. Its atrous convolution architecture provides capability to dissect complex street view imagery into discernible elements aligns well with the needs of urban planning and safety assessments.

\subsection{Machine Learning in Crime Prediction}
Recent advancements in machine learning have significantly enhanced crime prediction and analysis \cite{brantingham_randomized_2015}. Innovative methods, including linear regression, additive regression, and decision stump, have been applied to predict violent crime patterns, as shown in studies utilizing the Communities and Crime Unnormalized Dataset. Moreover, algorithms like K-nearest neighbor (KNN) and boosted decision trees have been implemented for analyzing extensive crime data, demonstrating the growing potential of machine learning in this field. Regarding urban greening, Lin applied spatial Durbin regressions to reveal the associations between street trees and crime in New York City. These developments indicate a promising direction for crime prediction and prevention, leveraging machine learning's capability to analyze complex patterns and trends in crime data.

\section{Method}
\subsection{Semantic Segmentation}
Semantic Segmentation Models are integral in advanced image analysis, particularly in understanding and interpreting complex scenes at a pixel level. These models, typically leveraging Convolutional Neural Networks (CNNs), are designed to classify each pixel in an image into predefined categories, providing a detailed delineation of the image content. In the context of our project, we have utilized a pre-trained semantic segmentation model, specifically the DeepLab V3 with a ResNet-101 backbone, to analyze street view images. The DeepLab V3 model, known for its efficiency in segmenting complex scenes, has been fine-tuned for our project using a Street View Dataset. Its advanced architecture, featuring atrous convolutions, is adept at capturing multi-scale context for improved feature extraction. 

\subsection{Regression Models}
To explore the relationship between urban street features and crime rates, we 
employed Linear Regression, Random Forest, and XGBoost on the collected dataset.

\subsubsection{Linear Regression}
The Linear Regression model is used as the baseline in our study. This model assumes a linear relationship between the dependent variable \( Y \) and one or more independent variables \( X_1, X_2, \ldots, X_n \). The model seeks to fit data by minimizing the sum of squared errors, and can be expressed as:
\[ Y = \beta_0 + \beta_1 X_1 + \beta_2 X_2 + \ldots + \beta_n X_n + \epsilon \]
where \( \beta_0, \beta_1, \ldots, \beta_n \) are the model coefficients, and \( \epsilon \) represents the error term. Despite its simplicity, Linear Regression provides a foundational understanding of relationships and serves as a benchmark for more complex models.

\subsubsection{Random Forest}
Random Forest is an ensemble learning method that operates by constructing a multitude of decision trees during training and outputs the average (for regression) or majority vote (for classification) of these trees. Specifically, each tree is built on a subset of the training data, with node splits based on a random subset of features. The Random Forest model can be represented as:
\[ Y = \frac{1}{B} \sum_{b=1}^{B} T_b(X; \Theta_b) \]
Here, \( B \) denotes the number of trees, \( T_b \) represents the prediction of the \( b \)-th tree, and \( \Theta_b \) is the vector of random parameters associated with this tree. This method is particularly effective in reducing overfitting by introducing randomness and aggregating multiple trees. It excels in handling large datasets and can manage both numerical and categorical features effectively.

\subsubsection{Decision Tree}
Decision Tree is a versatile machine learning technique used for predicting continuous numerical values. The model constructs a tree-like structure where each internal node represents a "test" on an attribute, each branch represents the outcome of the test, and each leaf node represents a value for the target variable. The paths from root to leaf represent classification rules. This approach is advantageous for its interpretability and ability to capture non-linear relationships.

\subsubsection{XGBoost}
XGBoost is an implementation of gradient boosted decision trees designed for speed and performance, which constructs its model by sequentially adding trees such that each subsequent tree aims to correct the errors made by the previous ones \cite{chen_xgboost_2016}. In its core operation, XGBoost minimizes the following objective function:
\[
\text{Obj} = \sum_{i=1}^n l(y_i, \hat{y}_i) + \sum_{k=1}^K \Omega(f_k),
\]
where \( l \) is the loss function measuring the difference between the predicted \( \hat{y}_i \) and the actual \( y_i \) values, \( K \) is the number of trees, \( f_k \) represents the \( k \)-th tree, and \( \Omega \) is the regularization term controlling the complexity of the model. XGBoost offers a robust solution to various types of predictive modeling problems, balancing accuracy and overfitting control.

\section{Evaluation Metrics}
To evaluate the performance of each model, we adopted two metrics: Mean Squared Error and R-squared (R²) Loss.

\subsection{Mean Squared Error (MSE)}
MSE reflects the average error of the model's predictions and is an important reference
for model selection and parameter tuning.

\subsection{R-squared (R²) Loss}
$R^2$ measures how closely the model's predicted values match the actual data points, i.e.,
the degree to which the model explains the variability in the data. A high $R^2$ value
indicates that the model can statistically explain the variations in crime rates well, which
is important for the accuracy of the predictions.

\subsection{Feature Importance}
In tree-based models such as Random Forest and XGBoost, it is possible to obtain quantified indicators of which features have the most impact on the predictions. This helps to identify and understand which street features are associated with higher crime rates.

\section{Preprocessing and Preliminary Experiments}
\subsection{Feature Selection}
Our analysis is structured around the community as the geographical unit. We source the demarcation of community boundaries from the NYC Open Data Platform. Given that each entry in both the crime records and the street view image dataset includes coordinate information, we assemble an extensive and current collection of street view imagery from multiple areas within New York City, employing the Google Street View API. Using GIS, we generate equidistant sampling points, spaced 50 meters apart, along NYC's roadways. Utilizing the GSV API, we acquire images at each sampling point to capture a panoramic representation of the streetscapes, primarily focusing on images captured in 2022. We also incorporate crime data from the same timeframe to ensure the relevance and timeliness of our analysis. 

After preprocessing, we aim to use the resulting 71 data points to train and evaluate our predictive model. In this model, crime rates serve as the target variable, while the features derived from street view images constitute the input variables.

\subsection{Preliminary Experiments}

In a preliminary experiment, we fine-tuned
the PyTorch implementation of the model, which was pre-trained on a subset of the COCO dataset \cite{lin_microsoft_2014}, on the Cityscapes dataset \cite{cordts_cityscapes_2016}. The Cityscapes dataset is a large-scale
benchmark collection tailored for the semantic understanding of urban street scenes. It contains a diverse array of stereo video sequences sourced from 50 different cities, captured at various times throughout the year.

Here, we utilized the gtFine training set from the Cityscapes dataset, which contains 2975 images, to generalize the model’s ability to segment a broader range of category data. We replaced the last layer of the classifier with a convolutional layer with a kernel size of 1x1 and a stride of 1x1.

Due to limitations on computational resources, we subsampled the training images from a size of 2048x1024 pixels to 512x512 pixels. We adopted a training scheme iterating through 40 epochs, with a batch size of 16.

\section{Experimental Analysis}
\subsection{Predictions Using Regression Models}
In this study, seven different regression models is trained to predict the aggregated crime rate of a community using randomly selected street view images. The models include: (1) Linear Regression, (2) Polynomial Regression, (3) Ridge Regression, (4) Support Vector Machine Regression, (5) Decision Tree Regression, (6) Random Forest Regression, and (7) Gradient Boosting Regression.

\subsubsection{Linear Regression}
For the baseline model, we linearly correlated the occurrence of each class of objects with the crime rate. The $R^2$ values of the baseline model is 0.02584, indicating the linear relationship between the occurrences of various object classes and the crime rate is weak, explaining only about 2.58\% of the variance in the crime rate.

\subsubsection{Polynomial Regression}
Two sets of experiment are carried out using Polynomial Regression models with degrees of 2 and 3. They all displayed poor performance, with negative R² values (-79074.65 and -1291838715.08, respectively) and high MSEs (6.99 and 114306.99).

\subsubsection{Ridge Regression}
In our Ridge regression experiment, we adjusted the $\alpha$ across five levels: 0.1, 1, 10, 100, and 1000, to optimize the prediction. As $\alpha$ increased, we observed an improvement in both metrics, suggesting a better model fit.

\subsubsection{Support Vector Machine}
In our experiment, we employed the Support Vector Regression (SVR) model, adjusting the epsilon parameter to observe its effect on model performance. Epsilon was varied across five values: 0.01, 0.1, 1, 10, and 100. 
\begin{figure}
    \centering
    \includegraphics[width=0.8\linewidth]{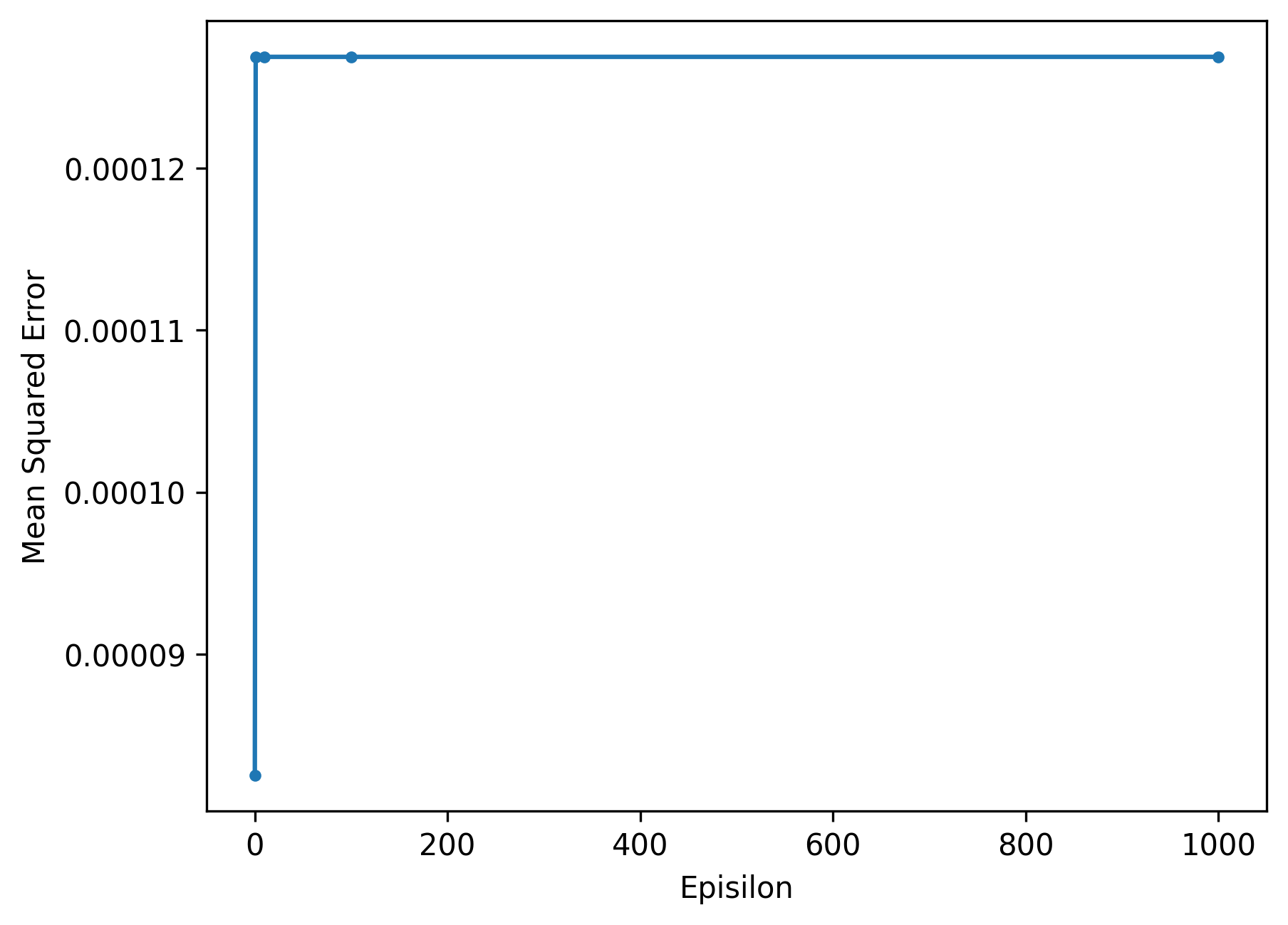}
    \caption{Mean Square Error in respect of $\epsilon$}
    \label{fig:svr}
\end{figure}
For an epsilon of 0.01, the model showed the best performance with an accuracy and R² value of 0.06737 and an MSE of 8.25e-05. However, as epsilon increased to 0.1, 1, 10, and 100, the model's performance drastically declined.

\subsubsection{Decision Tree}
In our experiment with the Decision Tree Regressor, we varied the max\_depth parameter across five levels: 2, 4, 6, 8, and 10. The model's performance was assessed using Accuracy, Mean Squared Error (MSE), and R-squared (R²). The best performance was observed at a depth of 4, with an accuracy and R² of 0.11769 and the lowest MSE of 7.81e-05. Increasing the depth beyond 4 resulted in a decrease in accuracy and R², and an increase in MSE, indicating potential overfitting. This suggests that a moderate depth provided the best balance between model complexity and predictive accuracy.

\subsubsection{Random Forest}
In our study, we experimented with the Random Forest Regressor, adjusting the number of estimators in the forest, which defines the number of trees in the forest. We tested a range of values: 10, 15, 20, 25, 30, 35, 40, 50, 100, 200, and 500. Across all tested number of estimators, the model consistently showed negative accuracy and R² values, with the least negative being at 500 trees (-0.09895 for both accuracy and R², and an MSE of 9.72e-05). As the number of trees increased, there was a general trend of improvement in these metrics.

\subsubsection{Gradient Boosting}
Similar to our findings with the Random Forest Regressor, we noticed a trend in the Gradient Boosting model's performance relative to the number of estimators.

This pattern indicates that while increasing the number of estimators initially improves the model's fit, there is a limit to this benefit. Exceeding this optimal number can lead to a plateau or even a decline in the model's predictive accuracy, emphasizing the need for a balanced approach to model complexity in machine learning.

\subsection{Feature Analysis}
We determine the feature importance for each tree model used in our analysis, with the rankings detailed in the subsequent tables. Intriguingly, across all three models, the “Aeroplane” feature emerged as the most influential in predicting crime rates. We hypothesize that this may be attributed to the vicinity of airports, which typically have heightened security measures and a diverse mix of individuals, potentially leading to more frequent detection and reporting of crimes. The significance of the second most important feature, “Cat", presents a more challenging aspect to interpret, and we leave its explanation as an open question for further investigation.

\begin{table}[ht]
\centering
\caption{Decision Tree Model (Depth=4)}
\label{tab:decision_tree_model}
\begin{tabular}{@{}ll@{}}
\toprule
Class & Importance \\
\midrule
Aeroplane & 0.74852879 \\
Cat & 0.12509578 \\
Person & 0.08090238 \\
Potted plant & 0.02141441 \\
Train & 0.01034135 \\
Car & 0.00864081 \\
TV/Monitor & 0.00507647 \\
... & ... \\
\bottomrule
\end{tabular}
\end{table}

\begin{table}[ht]
\centering
\caption{Random Forest Model}
\begin{tabular}{@{}ll@{}}
\toprule
Class & Importance \\
\midrule
Aeroplane & 0.463221813 \\
Cat & 0.339040096 \\
Potted plant & 0.119436065 \\
Car & 0.0274704138 \\
TV/Monitor & 0.0130055045 \\
Bird & 0.0110905086 \\
Person & 0.00727523173 \\
... & ... \\
\bottomrule
\end{tabular}
\end{table}

\begin{table}[ht]
\centering
\caption{Gradient Boosting Model}
\begin{tabular}{@{}ll@{}}
\toprule
Class & Importance \\
\midrule
Aeroplane & 0.691164777 \\
Cat & 0.142979583 \\
Potted plant & 0.107635157 \\
Sheep & 0.0193699097 \\
Car & 0.0135087689 \\
TV/Monitor & 0.0114400267 \\
Bird & 0.00415142709 \\
... & ... \\
\bottomrule
\end{tabular}
\end{table}

\subsection{Results Interpretation}
\begin{figure}
    \centering
    \includegraphics[width=0.8\linewidth]{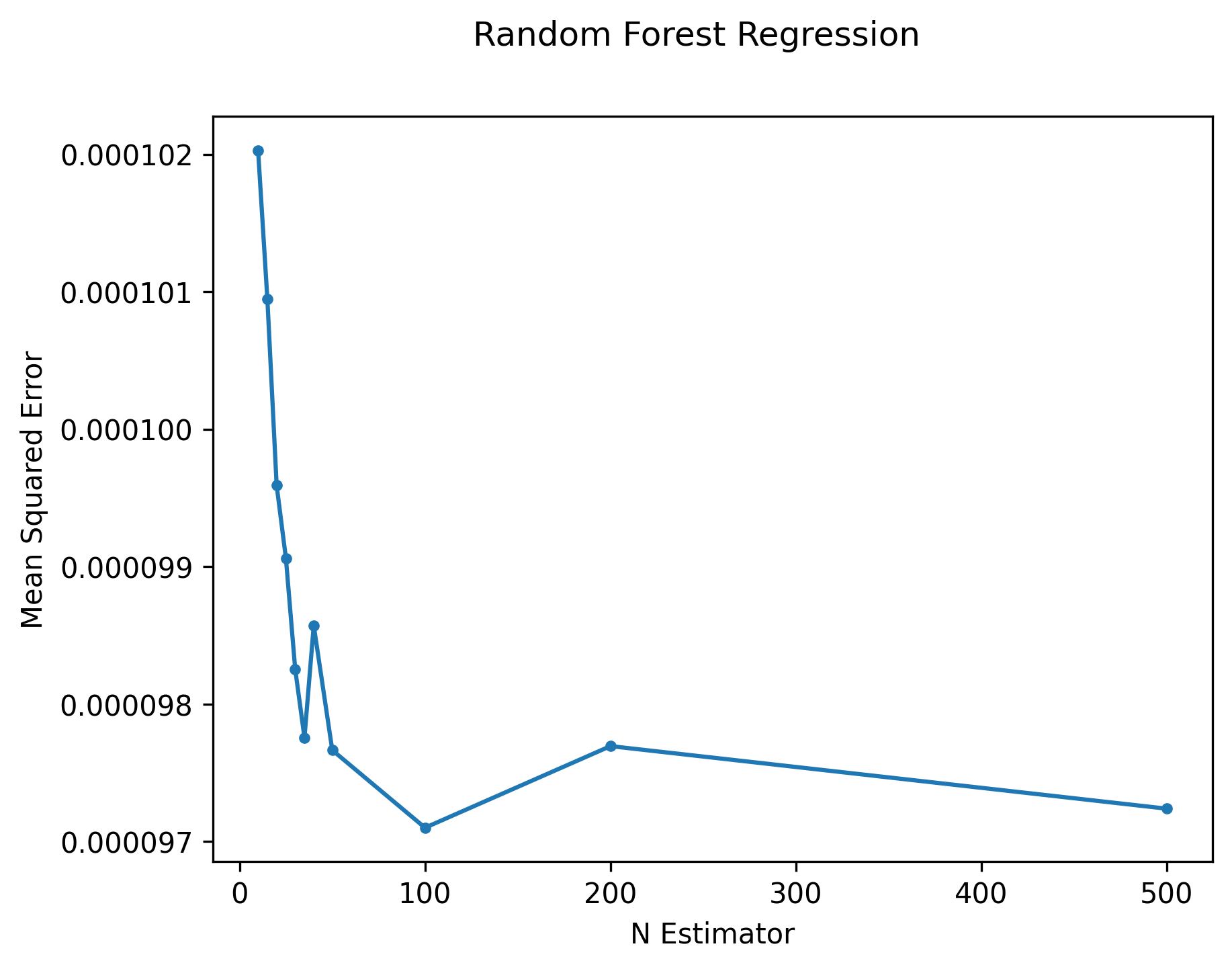}
    \caption{Random Forest Regression}
    \label{fig:random-forest}
\end{figure}

\begin{figure}[ht]
    \centering
    \includegraphics[width=0.8\linewidth]{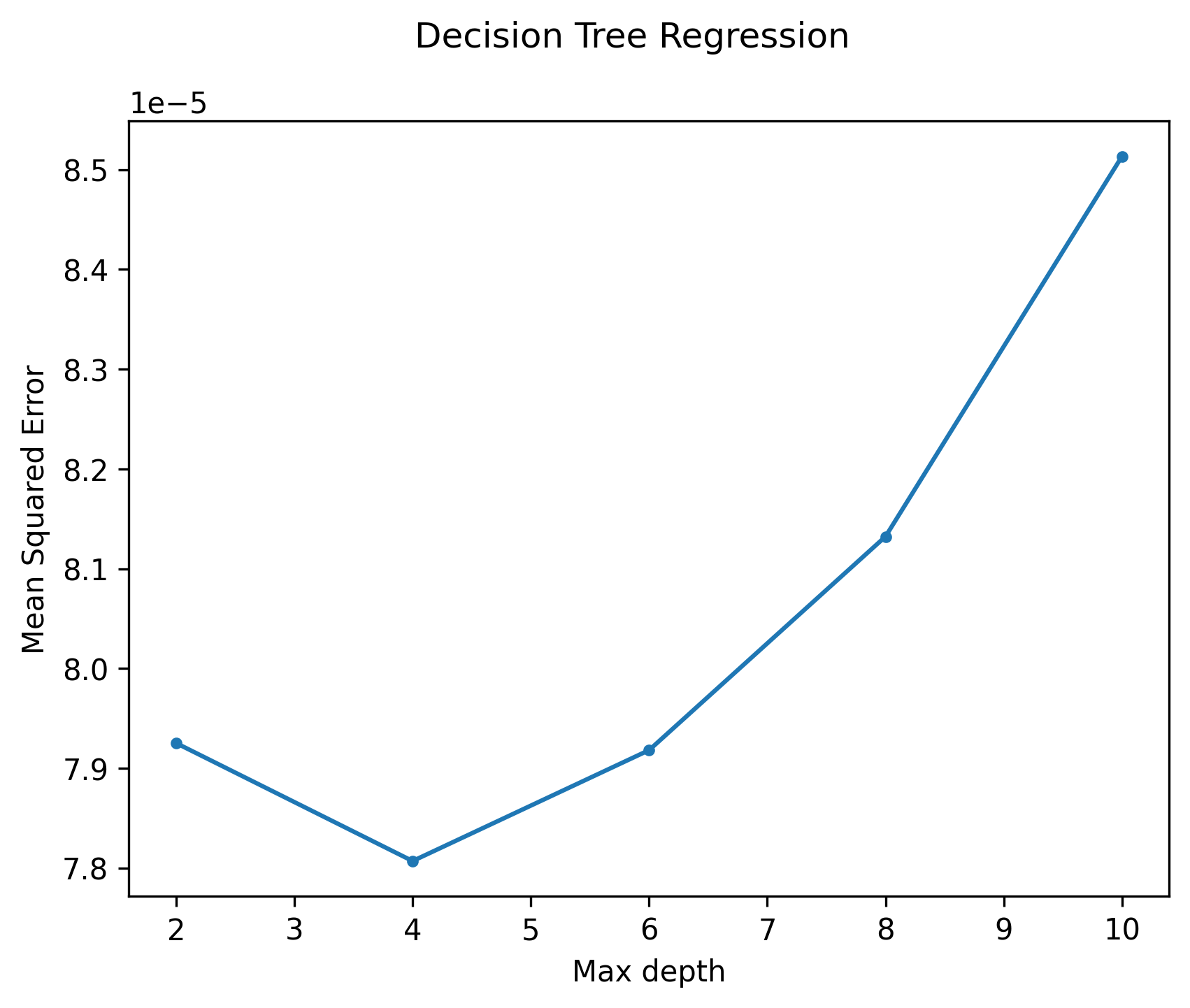}
    \caption{Decision Tree Regression}
    \label{fig:decision-tree}
\end{figure}

\begin{figure}[ht]
    \centering
    \includegraphics[width=0.8\linewidth]{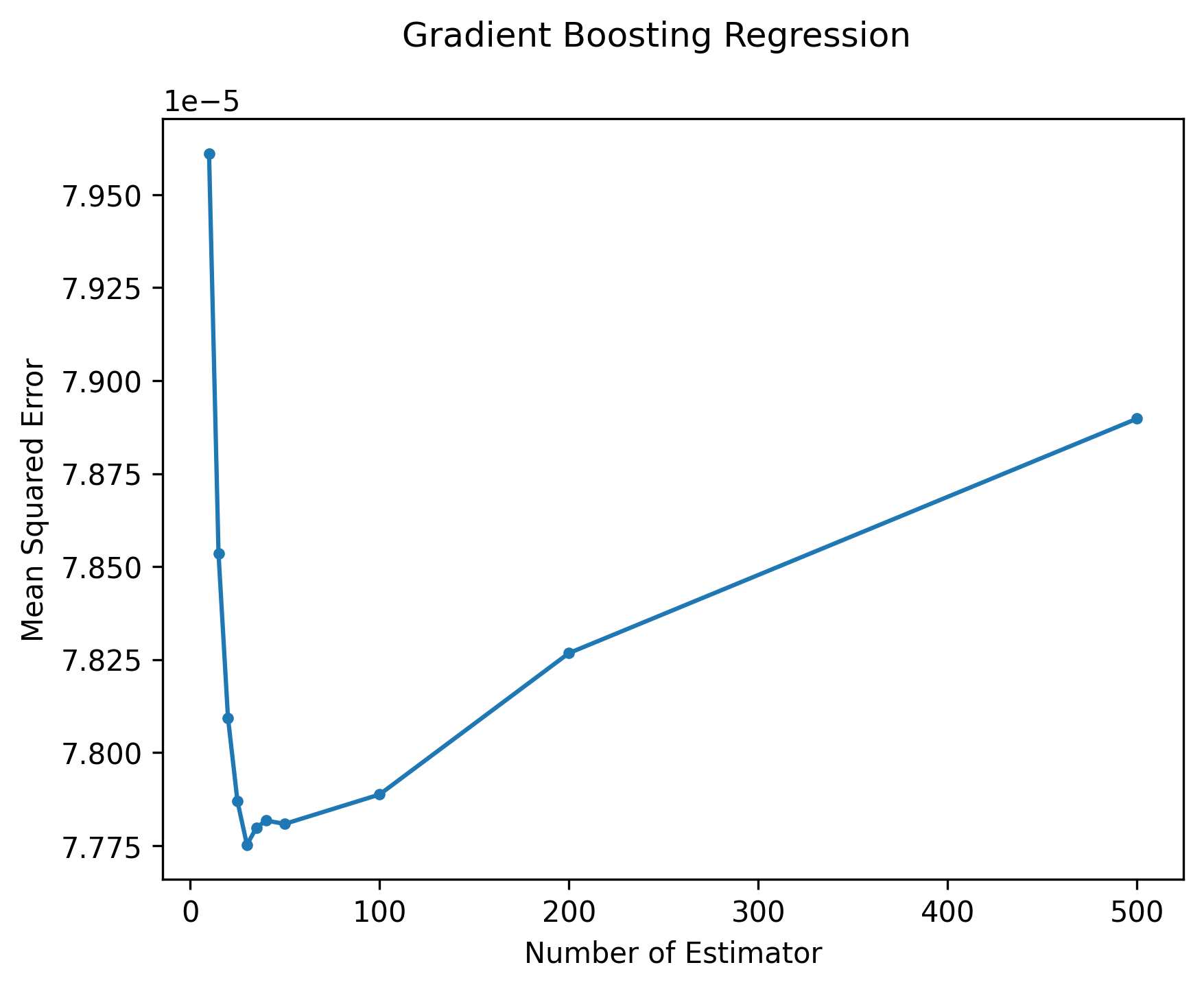}
    \caption{Gradient Boosting Regression}
    \label{fig:gradient-boosting}
\end{figure}

As observed from the variation in mean squared error (MSE) with the number of estimators, the decision tree model tends to exhibit overfitting, especially when the tree depth becomes excessively deep. This is a typical characteristic of decision trees, where increasing complexity can lead to models that are too closely fit to the training data, hence performing poorly on unseen data. In contrast, ensemble learning algorithms, such as Random Forest (RF) and Gradient Boosting Trees, have demonstrated a significant mitigation of this overfitting phenomenon. Notably, when comparing these two ensemble methods under similar conditions of tree quantity, Gradient Boosting Trees outperform RF in terms of accuracy. This superior performance is evident not only in specific scenarios but also in a global sense, as Gradient Boosting consistently achieves a higher best accuracy compared to RF.

\section{Conclusion and Future Work}
In this project, we utilized semantic segmentation models to analyze a comprehensive dataset of New York City street views, establishing a link between the visual elements of city streets and associated crime statistics. We specifically focused on evaluating how street view characteristics correlate with crime rates, types, and severity, and assessing the implications for urban intervention and crime prevention. Our findings, derived from employing Decision Tree Regression, Random Forest, and XGBoost models, indicate a significant correlation between certain urban features, notably the presence of aeroplanes, and crime rates. This unusual relationship suggests a higher incidence of crime in areas proximate to airports, potentially due to their relatively secluded locations and diverse population mix. Additionally, the concentration of security in these regions may lead to increased crime detection rates. These insights are pivotal for shaping future urban development policies and crime deterrence measures.

As we progress, our objective is to enhance the sophistication of our predictive models by integrating more advanced algorithms and a broader spectrum of variables. This enhancement aims to substantially increase both the accuracy and reliability of our predictions. Furthermore, we plan to extend our research to a more detailed scale, specifically at the level of census blocks. This refined approach is expected to yield a deeper understanding of the intricate dynamics between urban crime and its environmental factors. Additionally, the increase in data points at this granular level should provide our machine-learning algorithms with more robust datasets. We anticipate that this improvement in data richness will help address challenges such as overfitting, thereby refining the overall efficacy of our models.

While our findings are with limitations, they do suggest a novel and somewhat quirky note: be aware of airplanes and cats while navigating the streets of New York City to potentially enhance personal safety!

\bibliographystyle{aaai} 
\bibliography{references.bib}
\end{document}